\documentclass[11pt,a4paper]{article}
\usepackage[hyperref]{AACL-IJCNLP2020}
\usepackage{times}
\usepackage{latexsym}

\usepackage{bm}
\usepackage{graphicx}

\usepackage{colortbl}

\usepackage{microtype}

\usepackage{enumitem}
\setitemize{noitemsep,topsep=0pt,parsep=0pt,partopsep=0pt}

\aclfinalcopy 

\title{Measuring What Counts: \\The case of Rumour Stance Classification}

\author{Carolina Scarton \\ University of Sheffield, UK \\  \texttt{c.scarton@sheffield.ac.uk} \\
        \And  
        Diego Furtado Silva \\ Federal University of São Carlos, Brazil \\  \texttt{diegofs@ufscar.br} \\
        \AND
        Kalina Bontcheva \\ University of Sheffield, UK \\ \texttt{k.bontcheva.sheffield.ac.uk}}

\date{}

\begin{document}
\maketitle
\begin{abstract}
Stance classification can be a powerful tool for understanding whether and which users believe in online rumours.
The task aims to automatically predict the stance of replies towards a given rumour, namely \textit{support}, \textit{deny}, \textit{question}, or \textit{comment}. Numerous methods have been proposed and their performance compared in the RumourEval shared tasks in 2017 and 2019. Results demonstrated that this is a challenging problem since naturally occurring rumour stance data is highly imbalanced. This paper specifically questions the evaluation metrics used in these shared tasks.
We re-evaluate the systems submitted to the two RumourEval tasks and show that the two widely adopted metrics -- accuracy and macro-$F1$ -- are not robust for the four-class imbalanced task of rumour stance classification, as they wrongly favour systems with highly skewed accuracy towards the majority class. To overcome this problem, we propose new evaluation metrics for rumour stance detection. These are not only robust to imbalanced data but also score higher systems that are capable of recognising the two most informative minority classes (\textit{support} and \textit{deny}). 
\end{abstract}

\section{Introduction}

\begin{figure}[h]
\begin{center}
  \includegraphics[width=0.415\textwidth]{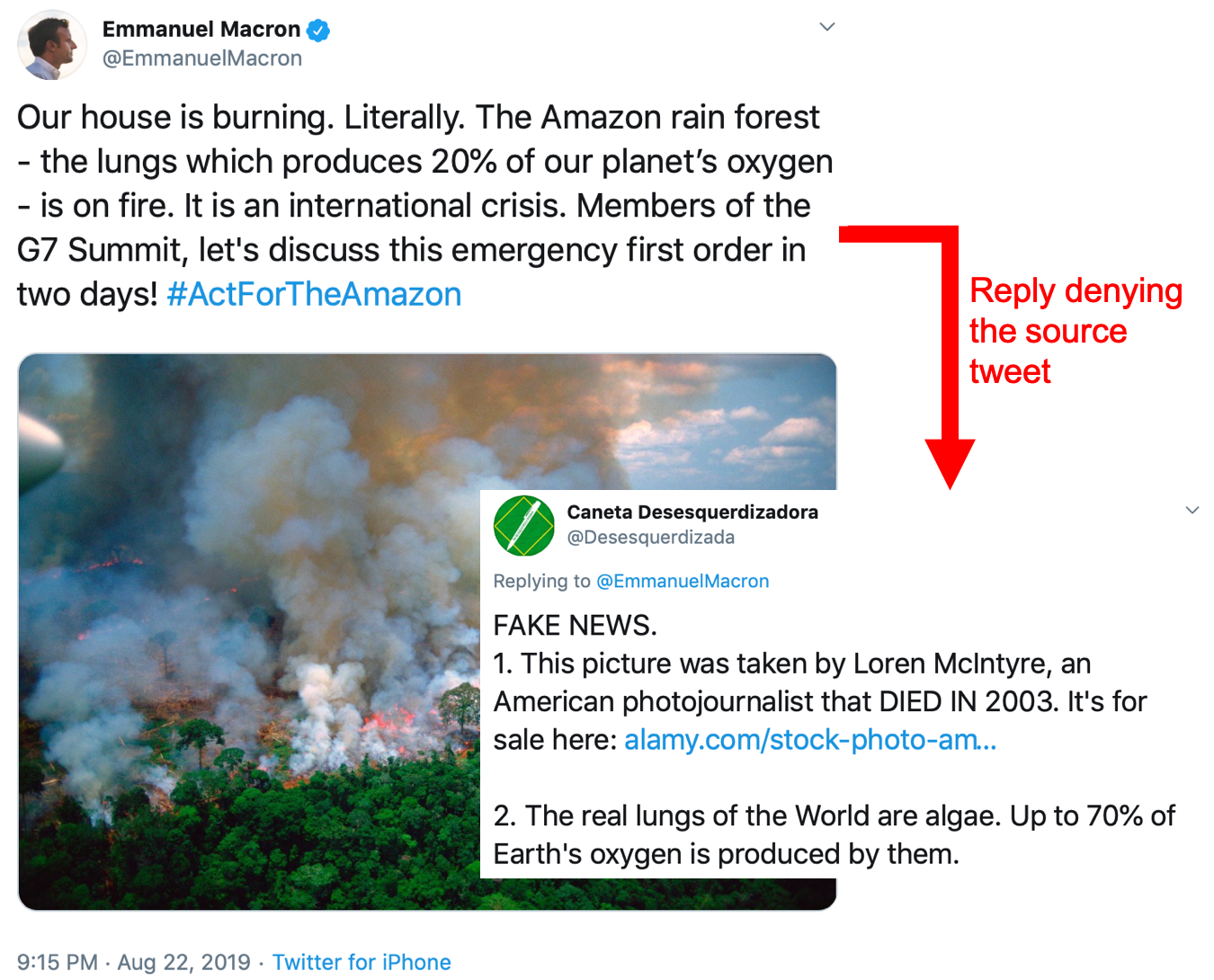}
  \caption{Example of a \textit{deny} stance.}
  \label{fig:ex1}
\end{center}
\vspace{-0.75em}
\end{figure}

The automatic analysis of online rumours has emerged as an important and challenging Natural Language Processing (NLP) task. Rumours in social media can be defined as claims that cannot be verified as true or false at the time of posting \cite{zubiaga-etal-2018}. Prior research \cite{mendoza2010twitter,kumar-carley-2019-tree} has shown that the stances of user replies are often a useful predictor of a rumour's likely veracity, specially in the case of false rumours that tend to receive a higher number of replies denying them \cite{zubiaga-etal-2016-plos}. 
However, their automatic classification is far from trivial as demonstrated by the results of two shared tasks -- RumourEval 2017 and 2019 \cite{derczynski-etal-2017-semeval,gorrell-etal-2019-semeval}. More specifically, sub-task A models rumour stance classification (RSC) as a four-class problem, where replies can:
\begin{itemize}
    \item \textbf{support}/agree with the rumour;
    \item \textbf{deny} the veracity of the rumour;
    \item \textbf{query}/ask for additional evidence;
    \item \textbf{comment} without clear contribution to assessing the veracity of the rumour.
\end{itemize}
\noindent{Figure \ref{fig:ex1} shows an example of a reply \textit{denying} a post on Twitter.} 

In RumourEval 2017 the training data contains 297 rumourous threads about eight events. The test set has 28 threads, with 20 threads about the same events as the training data and eight threads about unseen events. In 2019, the 2017 training data is augmented with 40 Reddit threads. The new 2019 test set has 56 threads about natural disasters from Twitter and a set of Reddit data (25 threads). 
These datasets for RSC are highly imbalanced: the \textit{comment} class is considerably larger than the other classes. Table \ref{tab:stats} shows the distribution of stances per class in both 2017 and 2019 datasets, where 66\% and 72\% of the data (respectively) correspond to \textit{comments}. \textit{Comments} arguably are the least useful when it comes to assessing overall rumour veracity, unlike \textit{support} and \textit{deny} which have been shown to help with rumour verification \cite{mendoza2010twitter}. Therefore, RSC is not only an imbalanced, multi-class problem, but it also has classes with different importance. This is different from standard stance classification tasks (e.g. SemEval 2016 task 6 \cite{mohammad-etal-2016-semeval}), where classes have arguably the same importance. It also differs from the veracity task (RumourEval sub-task B), where the problem is binary and it is not as an imbalanced problem as RSC.\footnote{Other NLP tasks, like sentiment analysis are also not comparable, since these tasks are either binary classification (which is then solved by using macro-$F1$) or do not have a clear priority over classes.} 

\begin{table}[t]
\centering
\scalebox{0.8}{
\begin{tabular}{l|cc}
\hline 
& \textbf{2017} & \textbf{2019} \\ \hline
support & 1,004 (18\%) & 1,184 (14\%) \\
deny & 415 (7\%) & 606 (7\%) \\
query & 464 (8\%) & 608 (7\%) \\
comment & 3,685 (66\%) & 6,176 (72\%) \\
\hline
total & 5,568 & 8,574 \\
\hline
\end{tabular}}
\caption{\label{tab:stats} Distribution of stances per class -- with percentages between parenthesis.}
\end{table}

RumourEval 2017 evaluated systems based on accuracy ($ACC$), which is not sufficiently robust on imbalanced datasets \cite{huang2005using}. This prompted the adoption of macro-$F1$ in the 2019 evaluation. \newcite{kumar-carley-2019-tree} also argue that macro-$F1$ is a more reliable evaluation metric for RSC. Previous work on RSC also adopted these metrics \cite{li-etal-2019-rumor,kochkina-etal-2018-one,dungs-etal-2018-rumour}.

This paper re-evaluates the sub-task A results of RumourEval 2017 and 2019.\footnote{We thank the organisers for making the data available.} It analyses the performance of the participating systems according to different evaluation metrics and 
shows that even macro-$F1$, that is robust for evaluating binary classification on imbalanced datasets, fails to reliably evaluate the performance on RSC. This is particularly critical in RumourEval where not only is data imbalanced, but also two minority classes (\textit{deny} and \textit{support}) are the most important to classify well. Based on prior research on imbalanced datasets in areas other that NLP (e.g. \newcite{yijing-etal-2016} and \newcite{elrahman-and-abraham2013}), we propose four alternative metrics for evaluating RSC. These metrics change the systems ranking for RSC in RumourEval 2017 and 2019, rewarding systems with high performance on the minority classes. 

\section{Evaluation metrics for classification}\label{sec:metrics}

We define $TP = $ true positives, $TN = $ true negatives, $FP = $ false positives and $FN =$ false negatives, where $TP_c$ ($FP_c$) is equivalent to the true (false) positives and $TN_c$ ($FN_c$) is equivalent to the true (false) negatives for a given class $c$.
\paragraph{Accuracy ($\bm{ACC}$)} is the ratio between the number of correct predictions and the total number of predictions ($N$): $ACC =\frac{\sum_{c=1}^{C} TP_c}{N}$, where $C$ is the number of classes. $ACC$ only considers the values that were classified correctly, disregarding the mistakes.
This is inadequate for imbalanced problems like RSC where, as shown in Table \ref{tab:stats}, most of the data is classified as \textit{comments}. 
As shown in Section~\ref{sec:results}, most systems will fail to classify the \textit{deny} class and still achieve high scores in terms of $ACC$. In fact, the best system for 2017 according to $ACC$ (\texttt{Turing}) fails to classify all \textit{denies}. 



\paragraph{Precision ($\bm P_c$) and Recall ($\bm R_c$)} $P_c$ is the ratio between the number of correctly predicted instances and all the predicted values for $c$: $P_c=\frac{TP_c}{TP_c + FP_c}$. $R_c$ is the ratio between correctly predicted instances and the number of instances that actually belongs to the class $c$: $R_c=\frac{TP_c}{TP_c + FN_c}$.

\paragraph{macro-$\bm{F\beta}$} 
$F\beta_c$ score is defined as the harmonic mean of precision and recall, where the per-class score can be defined as: $F{\beta}_c = (1 + \beta^2) \frac{P_c \cdot R_c}{\beta^2 P_c + R_c}$.
If $\beta = 1$, $F{\beta}$ is the $F1$ score. If $\beta > 1$, $R$ is given a higher weight and if $\beta < 1$, $P$ is given a higher weight.
The macro-$F{\beta}$ is the arithmetic mean between the $F{\beta}$ scores for each class: macro-$F{\beta}_c = \frac{\sum_{c=1}^{C}F{\beta}_c}{C}$.
Although macro-$F1$ is expected to perform better than $ACC$ for imbalanced binary problems, its benefits in the scenario of multi-class classification are not clear. Specifically, as it relies on the arithmetic mean over the classes, it may hide the poor performance of a model in one of the classes if it performs well on the majority class (i.e. \textit{comments} in this case). For instance, as shown in Table \ref{tab:results2017}, according to macro-$F1$ the best performing system would be \textit{ECNU}, which still fails to classify correctly almost all \textit{deny} instances.

\begin{table*}[!t]
\centering
\scalebox{0.7}{
\begin{tabular}{l|>{\columncolor[gray]{0.9}}c|>{\columncolor[gray]{0.9}}c||c|c|c|c}
\hline 
& $\bm{ACC}$ & macro-$F1$ & $GMR$ & $wAUC$ & $wF1$ & $wF2$ \\ \hline
\texttt{Turing} a & \textbf{0.784} (1) & 0.434 (5) & 0.000 (8) & 0.583 (7) & 0.274 (6) & 0.230 (7)   \\
\texttt{UWaterloo} \cite{bahuleyan-vechtomova-2017-uwaterloo} & 0.780 (2) & 0.455 (2) & 0.237 (5) & 0.595 (5) & 0.300 (2) & 0.255 (6)  \\
\texttt{ECNU} \cite{wang-etal-2017-ecnu} & 0.778 (3) & \textbf{0.467} (1) & 0.214 (7) & 0.599 (4) & 0.289 (4) & 0.263 (4)  \\
\texttt{Mama Edha} \cite{garcia-lozano-etal-2017-mama} & 0.749 (4) & 0.453 (3) & 0.220 (6) & \textbf{0.607} (1) & 0.299 (3) & 0.283 (3) \\
\texttt{NileTMRG} \cite{enayet-el-beltagy-2017-niletmrg} & 0.709 (5) & 0.452 (4) & \textbf{0.363} (1) & 0.606 (2) & \textbf{0.306} (1) & \textbf{0.296} (1)  \\
\texttt{IKM} \cite{chen-etal-2017-ikm} & 0.701 (6) & 0.408 (7) & 0.272 (4) & 0.570 (8) & 0.241 (7) & 0.226 (8) \\
\texttt{IITP} \cite{singh-etal-2017-iitp} & 0.641 (7) & 0.403 (8) & 0.345 (2) & 0.602 (3) & 0.276 (5) & 0.294 (2) \\
\texttt{DFKI DKT} \cite{srivastava-etal-2017-dfki} & 0.635 (8) & 0.409 (6) & 0.316 (3) & 0.589 (6)  & 0.234 (8) & 0.256 (5)  \\
\hline
\hline
\texttt{majority class} & 0.742 & 0.213 & 0.000 & 0.500 & 0.043 & 0.047   \\
\texttt{all denies} & 0.068 & 0.032 & 0.000 & 0.500 & 0.051 & 0.107  \\
\texttt{all support} & 0.090 & 0.041 & 0.000 & 0.500 & 0.066 & 0.132 \\
\hline
\end{tabular}}
\caption{\label{tab:results2017} Evaluation of RumourEval 2017 submissions. Values between parenthesis are the ranking of the system according to the metric. The official evaluation metric column ($ACC$) is highlighted in bold.}
\end{table*}

\begin{table*}[!t]
\centering
\scalebox{0.7}{
\begin{tabular}{l|>{\columncolor[gray]{0.9}}c|>{\columncolor[gray]{0.9}}c||c|c|c|c}
\hline 
& $ACC$ & \textbf{macro-}$\bm{F1}$ & $GMR$ & $wAUC$ & $wF1$ & $wF2$   \\ \hline
\texttt{BLCU NLP} \cite{yang-etal-2019-blcu} & 0.841 (2) & \textbf{0.619} (1) & 0.571 (2) & 0.722 (2) & \textbf{0.520} (1) & 0.500 (2)\\
\texttt{BUT-FIT} \cite{fajcik-etal-2019-fit} & \textbf{0.852} (1) & 0.607 (2) & 0.519 (3) & 0.689 (3) & 0.492 (3) & 0.441 (3) \\
\texttt{eventAI} \cite{li-etal-2019-eventai} & 0.735 (11) & 0.578 (3) & \textbf{0.726} (1) & \textbf{0.807} (1) & 0.502 (2) & \textbf{0.602} (1)  \\
\texttt{UPV} \cite{ghanem-etal-2019-upv} & 0.832 (4) & 0.490 (4) & 0.333 (5) & 0.614 (5) & 0.340 (4) & 0.292 (5) \\
\texttt{GWU} \cite{hamidian-diab-2019-gwu} & 0.797 (9) & 0.435 (5) & 0.000 (7) & 0.604 (6)  & 0.284 (5) & 0.265 (6) \\
\texttt{SINAI-DL} \cite{garcia-cumbreras-etal-2019-sinai} & 0.830 (5) & 0.430 (6) & 0.000 (8) & 0.577 (7) & 0.255 (7) & 0.215 (7) \\
\texttt{wshuyi} & 0.538 (13) & 0.370 (7) & 0.467 (4) & 0.627 (4) & 0.261 (6) & 0.325 (4)  \\
\texttt{Columbia} \cite{liu-etal-2019-columbia} & 0.789 (10) & 0.363 (8) & 0.000 (9) & 0.562 (10) & 0.221 (10) & 0.191 (9) \\
\texttt{jurebb} & 0.806 (8) & 0.354 (9) & 0.122 (6) & 0.567 (9) & 0.229 (8) & 0.120 (12) \\
\texttt{mukundyr} & 0.837 (3) & 0.340 (10) & 0.000 (10) & 0.570 (8) & 0.224 (9) & 0.198 (8) \\
\texttt{nx1} & 0.828 (7) & 0.327 (11) & 0.000 (11) & 0.557 (11) & 0.206 (11) & 0.173 (10) \\
\texttt{WeST} \cite{baris-etal-2019-clearumor} & 0.829 (6) & 0.321 (12) & 0.000 (12) & 0.551 (12) & 0.197 (12) & 0.161 (11) \\
\texttt{Xinthl} & 0.725 (12) & 0.230 (13) & 0.000 (13) & 0.493 (13) & 0.072 (13) & 0.071 (13) \\
\hline
\hline
\texttt{majority class} & 0.808 & 0.223 & 0.000 & 0.500 & 0.045 & 0.048  \\
\texttt{all denies} & 0.055 & 0.026 & 0.000 & 0.500 & 0.042 & 0.091\\
\texttt{all support} & 0.086 & 0.040 & 0.000 & 0.500 & 0.063 & 0.128\\
\hline
\end{tabular}}
\caption{\label{tab:results2019} Evaluation of RumourEval 2019 submissions. Values between parenthesis are the ranking of the system according to the metric. The official evaluation metric column (macro-$F1$) is highlighted in bold.}
\end{table*}

\begin{figure*}[ht]
\begin{center}
  \includegraphics[width=.95\textwidth]{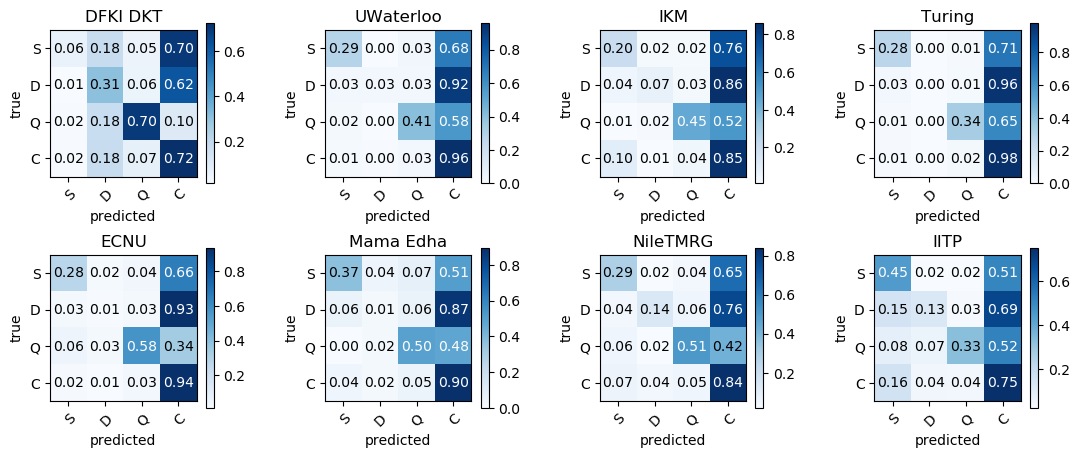}
  \caption{Confusion matrix for systems from RumourEval 2017.}
  \label{fig:cm2017}
\end{center}
\end{figure*}


\begin{figure*}[!t]
\begin{center}
  \includegraphics[width=.95\textwidth]{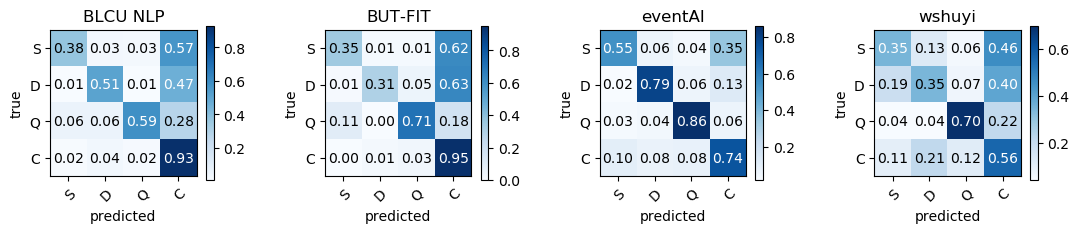}
  \caption{Confusion matrix for selected systems from RumourEval 2019. All other systems failed to classify correctly either all or the vast majority of \textit{deny} instances.}
  \label{fig:cm2019}
\end{center}
\end{figure*}

\paragraph{Geometric mean} Metrics like the geometric mean of $R$: 

\vspace{-0.6em} 
\[
GMR = \sqrt[C]{\prod_{c=1}^{C} R_c}.
\]
\vspace{-0.7em} 

\noindent{are proposed for evaluating specific types of errors. As $FNs$ may be more relevant than $FPs$ for imbalanced data, assessing models using $R$ is an option to measure this specific type of error. Moreover, applying $GMR$ for each class severely penalises a model that achieves a low score for a given class.}

\paragraph{Area under the $\bm{ROC}$ curve} Receiver operating characteristic ($ROC$) \cite{fawcett2006roc} assesses the performance of classifiers considering the relation between $R_c$ and the false positive rate, defined as (per class): $FPR_c=\frac{FP_c}{TN_c + FP_c}$.
Since RSC consists of discrete classifications, $ROC$ charts for each $c$ contain only one point regarding the coordinate ($FPR_c$, $R_c$). 
Area under the $ROC$ curve ($AUC$) measures the area of the curve produced by the points in an $ROC$ space. In the discrete case, it measures the area of the polygon drawn by the segments connecting the vertices $((0,0), (FPR_c,R_c), (1,1), (0,1))$. 
High $AUC$ scores are achieved when $R$ (probability of detection) is maximised, while $FPR$ (probability of false alarm) is minimised.
We experiment with a weighted variation of $AUC$: 
\vspace{-0.7em} 
\[
wAUC = \sum_{c=1}^{C}w_c \cdot AUC_c.
\]
\vspace{-0.7em} 


\paragraph{Weighted macro-$\bm{F\beta}$} a variation of macro-$F{\beta}$, where each class also receives different weights, is also considered: 

\vspace{-0.7em} 
\[
wF{\beta} = \sum_{c=1}^{C}w_c \cdot F{\beta}_c,
\]
\vspace{-0.7em} 

\noindent We use $\beta = 1$ ($P$ and $R$ have the same importance) and $\beta = 2$ ($R$ is more important). 
Arguably, misclassifying \textit{denies} and \textit{supports} ($FN_{D}$ and $FN_{S}$, respectively) is equivalent to ignore relevant information for debunking a rumour. Since $FNs$ negatively impact $R$, we hypothesise that $\beta = 2$ is more robust for the RSC case.  

$wAUC$ and $wF{\beta}$ are inspired by empirical evidence that different classes have different importance for RSC.\footnote{Similarly, previous work proposes metrics \cite{elkan2001foundations} and learning algorithms \cite{chawla2008automatically} based on class-specific mis-classification costs.} Weights should be manually defined, since they cannot be automatically learnt. We follow the hypothesis that \textit{support} and \textit{deny} classes are more informative than others.\footnote{$w_{support} = w_{deny} = 0.40$, $w_{query} = 0.15$ and $w_{comment} = 0.05$.}



\section{Re-evaluating RumourEval task A} \label{sec:results}

Tables \ref{tab:results2017} and \ref{tab:results2019} report the different evaluation scores per metric for each of the RumourEval 2017 and 2019 systems.\footnote{The systems \texttt{HLT(HITSZ)}, \texttt{LECS}, \texttt{magc}, \texttt{UI-AI}, \texttt{shaheyu} and \texttt{NimbusTwoThousand} are omitted because they do not provide the same number of inputs as the test set.} 
$ACC$ and macro-$F1$ are reported in the second and third columns respectively, followed by a column for each of the four proposed metrics. 
Besides evaluating the participating systems, we also computed scores for three baselines: \texttt{majority class} (all stances are considered \textit{comments}), \texttt{all denies} and \texttt{all support} (all replies are classed as \textit{deny}/\textit{support}).

Our results show that the choice of evaluation metric has a significant impact on system ranking. In RumourEval 2017, the winning system based on $ACC$ was \texttt{Turing}. 
However, Figure \ref{fig:cm2017} shows that this system classified all \textit{denies} incorrectly, favouring the majority class (\textit{comment}). When looking at the macro-$F1$ score, \texttt{Turing} is classified as fifth, whilst the winner is \texttt{ECNU}, followed by \texttt{UWaterloo}. Both systems also perform very poorly on \textit{denies}, classifying only 1\% and 3\% of them correctly. On the other hand, the four proposed metrics penalise these systems for these errors and rank higher those that perform better on classes other than the majority one. For example, the winner according to $GMR$, $wF1$ and $wF2$ is \texttt{NileTMRG} that, according to Figure \ref{fig:cm2017}, shows higher accuracy on the \textit{deny}, \textit{support} and \textit{query} classes, without considerably degraded performance on the majority class. $wAUC$ still favours the \texttt{Mama Edha} system which has very limited performance on the important \textit{deny} class. As is evident from Figure \ref{fig:cm2017}, \texttt{NileTMRG} is arguably the best system in predicting all classes: it has the highest accuracy for \textit{denies}, and a sufficiently high accuracy for \textit{support}, \textit{queries} and \textit{comments}. Using the same criteria, the second best system should be \texttt{IITP}. The only two metrics that reflect this ranking are $GMR$ and $wF2$. In the case of $wF1$, the second system is \texttt{UWaterloo}, which has a very low accuracy on the \textit{deny} class.

For RumourEval 2019, the best system according to macro-$F1$ (the official metric) is \texttt{BLCU NLP}, followed by \texttt{BUT-FIT}. However, after analysing the confusion matrices in Figure \ref{fig:cm2019}, we can conclude that \texttt{eventAI} is a more suitable model due to its high accuracy on \textit{support} and  \textit{deny}. Metrics $GMR$, $wAUC$ and $wF2$ show \texttt{eventAI} as the best system. Finally, \texttt{wshuyi} is ranked as fourth according to $GMR$, $wAUC$ and $wF2$, while it ranked seventh in terms of macro-$F1$, behind systems like \texttt{GWU} and \texttt{SINAI-DL} that fail to classify all \textit{deny} instances. Although \texttt{wshuyi} is clearly worse than \texttt{eventAI}, \texttt{BLCU NLP} and \texttt{BUT-FIT}, it is arguably more reliable than systems that misclassify the large majority of \textit{denies}.\footnote{Confusion matrices for all systems of RumourEval 2019 are presented in Appendix \ref{sec:appendix}.} Our analyses suggest that $GMR$ and $wF2$ are the most reliable for evaluating RSC tasks. 



\section{Weight selection}

In Section \ref{sec:results}, $wAUC$, $wF1$ and $wF2$ have been obtained using empirically defined weights ($w_{support} = w_{deny} = 0.40$, $w_{query} = 0.15$ and $w_{comment} = 0.05$). These values reflect the key importance of the \textit{support} and \textit{deny} classes. Although \textit{query} is less important than the first two, it is nevertheless more informative than \textit{comment}.

Previous work tried to adjust the learning weights in order to minimise the effect of the imbalanced data. \newcite{garcia-lozano-etal-2017-mama} (\texttt{Mama Edha}), change the weights of their Convolutional Neural Network (CNN) architecture, giving higher importance to \textit{support}, \textit{deny} and \textit{query} classes, to better reflect their class distribution.\footnote{$w_{support} = 0.157, w_{deny} = 0.396$, $w_{query} = 0.399$ and $w_{comment} = 0.048$} \newcite{ghanem-etal-2019-upv} (\texttt{UPV}) also change the weights in their Logistic Regression model in accordance with the data distribution criterion.\footnote{$w_{support} = 0.2, w_{deny} = 0.35$, $w_{query} = 0.35$ and $w_{comment} = 0.1$} Nevertheless, these systems misclassify almost all \textit{deny} instances.

Table \ref{tab:results2017_weights} shows the RumourEval 2017 systems ranked according to $wF2$ using the \texttt{Mama Edha} and \texttt{UPV} weights. In these cases, $wF2$ benefits \texttt{DFKI DKT}, ranking it first, since \textit{queries} receive a higher weight than \textit{support}. However, this system only correctly classifies 6\% of \textit{support} instances, which makes it less suitable for our task than \texttt{NileTMRG} for instance. \texttt{ECNU} is also ranked better than \texttt{Mama Edha} and \texttt{IITP}, likely due to its higher performance on \textit{query} instances.

\begin{table}[!ht]
\centering
\scalebox{0.7}{
\begin{tabular}{l|c|c}
\hline 
& $wF2$ & $wF2$\\ 
& \texttt{Mama Edha} & \texttt{UPV} \\\hline
\texttt{Turing} & 0.246 (8) & 0.289 (8)  \\
\texttt{UWaterloo} & 0.283 (7) & 0.322 (5)  \\
\texttt{ECNU} & 0.334 (3) & 0.364 (3) \\
\texttt{Mama Edha} & 0.312 (4) & 0.349 (4) \\
\texttt{NileTMRG} & 0.350 (2) & 0.374 (2) \\
\texttt{IKM} & 0.293 (5) & 0.318 (7) \\
\texttt{IITP} & 0.289 (6) & 0.321 (6) \\
\texttt{DFKI DKT} & \textbf{0.399} (1) & \textbf{0.398} (1)  \\
\hline
\end{tabular}}
\caption{\label{tab:results2017_weights} RumourEval 2017 evaluated using $wF2$ with weights from \texttt{Mama Edha} and \texttt{UPV}.}
\end{table}

Arguably, defining weights based purely on data distribution is not sufficient for RSC. Thus our empirically defined weights seem to be more suitable than those derived from data distribution alone, as the former accurately reflect that \textit{support} and \textit{deny} are the most important, albeit minority distributed classes. Further research is required in order to identify the most suitable weights for this task.

\section{Discussion}

This paper re-evaluated the systems that participated in the two editions of RumourEval task A (stance classification). We showed that the choice of evaluation metric for assessing the task has a significant impact on system rankings.
The metrics proposed here are better suited to evaluating tasks with imbalanced data, since they do not favour the majority class. We also suggest variations of $AUC$ and macro-$F\beta$ that give different weights for each class, which is desirable for scenarios where some classes are more important than others. 

The main lesson from this paper is that evaluation is an important aspect of NLP tasks and it needs to be done accordingly, after a careful consideration of the problem and the data available. In particular, we recommend that future work on RSC uses $GMR$ and/or $wF\beta$ (preferably $\beta = 2$) as evaluation metrics. Best practices on evaluation rely on several metrics that can assess different aspects of quality. Therefore, relying on several metrics is likely the best approach for RSC evaluation.

\section*{Acknowledgments}
This work was funded by the WeVerify project (EU H2020, grant agreement: 825297). The SoBigData TransNational Access program (EU H2020, grant agreement: 654024) funded Diego Silva's visit to the University of Sheffield.


\bibliography{anthology,acl2020}

\begin{thebibliography}{31}
\expandafter\ifx\csname natexlab\endcsname\relax\def\natexlab#1{#1}\fi

\bibitem[{Bahuleyan and Vechtomova(2017)}]{bahuleyan-vechtomova-2017-uwaterloo}
Hareesh Bahuleyan and Olga Vechtomova. 2017.
\newblock \href {https://doi.org/10.18653/v1/S17-2080} {{UW}aterloo at
  {S}em{E}val-2017 task 8: Detecting stance towards rumours with topic
  independent features}.
\newblock In \emph{Proceedings of the 11th International Workshop on Semantic
  Evaluation ({S}em{E}val-2017)}, pages 461--464, Vancouver, Canada.
  Association for Computational Linguistics.

\bibitem[{Baris et~al.(2019)Baris, Schmelzeisen, and
  Staab}]{baris-etal-2019-clearumor}
Ipek Baris, Lukas Schmelzeisen, and Steffen Staab. 2019.
\newblock \href {https://doi.org/10.18653/v1/S19-2193} {{CLEAR}umor at
  {S}em{E}val-2019 task 7: {C}onvo{L}ving {ELM}o against rumors}.
\newblock In \emph{Proceedings of the 13th International Workshop on Semantic
  Evaluation}, pages 1105--1109, Minneapolis, Minnesota, USA. Association for
  Computational Linguistics.

\bibitem[{Chawla et~al.(2008)Chawla, Cieslak, Hall, and
  Joshi}]{chawla2008automatically}
Nitesh~V. Chawla, David~A. Cieslak, Lawrence~O. Hall, and Ajay Joshi. 2008.
\newblock \href {https://doi.org/10.1007/s10618-008-0087-0} {Automatically
  countering imbalance and its empirical relationship to cost}.
\newblock \emph{Data Mining and Knowledge Discovery}, 17(2):225--252.

\bibitem[{Chen et~al.(2017)Chen, Liu, and Kao}]{chen-etal-2017-ikm}
Yi-Chin Chen, Zhao-Yang Liu, and Hung-Yu Kao. 2017.
\newblock \href {https://doi.org/10.18653/v1/S17-2081} {{IKM} at
  {S}em{E}val-2017 task 8: Convolutional neural networks for stance detection
  and rumor verification}.
\newblock In \emph{Proceedings of the 11th International Workshop on Semantic
  Evaluation ({S}em{E}val-2017)}, pages 465--469, Vancouver, Canada.
  Association for Computational Linguistics.

\bibitem[{Derczynski et~al.(2017)Derczynski, Bontcheva, Liakata, Procter, Wong
  Sak~Hoi, and Zubiaga}]{derczynski-etal-2017-semeval}
Leon Derczynski, Kalina Bontcheva, Maria Liakata, Rob Procter, Geraldine Wong
  Sak~Hoi, and Arkaitz Zubiaga. 2017.
\newblock \href {https://doi.org/10.18653/v1/S17-2006} {{S}em{E}val-2017 task
  8: {R}umour{E}val: Determining rumour veracity and support for rumours}.
\newblock In \emph{Proceedings of the 11th International Workshop on Semantic
  Evaluation ({S}em{E}val-2017)}, pages 69--76, Vancouver, Canada. Association
  for Computational Linguistics.

\bibitem[{Dungs et~al.(2018)Dungs, Aker, Fuhr, and
  Bontcheva}]{dungs-etal-2018-rumour}
Sebastian Dungs, Ahmet Aker, Norbert Fuhr, and Kalina Bontcheva. 2018.
\newblock \href {https://www.aclweb.org/anthology/C18-1284} {Can rumour stance
  alone predict veracity?}
\newblock In \emph{Proceedings of the 27th International Conference on
  Computational Linguistics}, pages 3360--3370, Santa Fe, New Mexico, USA.
  Association for Computational Linguistics.

\bibitem[{Elkan(2001)}]{elkan2001foundations}
Charles Elkan. 2001.
\newblock \href {https://www.ijcai.org/Proceedings/01/IJCAI-2001-k.pdf} {The
  foundations of cost-sensitive learning}.
\newblock In \emph{Proceedings of the 17th International Joint Conference on
  Artificial Intelligence - Volume 2}, pages 973--978, Seattle, Washington,
  USA. International Joint Conferences on Artificial Intelligence.

\bibitem[{Elrahman and Abraham(2013)}]{elrahman-and-abraham2013}
Shaza M.~Abd Elrahman and Ajith Abraham. 2013.
\newblock \href {http://www.softcomputing.net/jnic2.pdf} {{A Review of Class
  Imbalance Problem}}.
\newblock \emph{Journal of Network and Innovative Computing}, 1:332--340.

\bibitem[{Enayet and El-Beltagy(2017)}]{enayet-el-beltagy-2017-niletmrg}
Omar Enayet and Samhaa~R. El-Beltagy. 2017.
\newblock \href {https://doi.org/10.18653/v1/S17-2082} {{N}ile{TMRG} at
  {S}em{E}val-2017 task 8: Determining rumour and veracity support for rumours
  on twitter.}
\newblock In \emph{Proceedings of the 11th International Workshop on Semantic
  Evaluation ({S}em{E}val-2017)}, pages 470--474, Vancouver, Canada.
  Association for Computational Linguistics.

\bibitem[{Fajcik et~al.(2019)Fajcik, Smrz, and Burget}]{fajcik-etal-2019-fit}
Martin Fajcik, Pavel Smrz, and Lukas Burget. 2019.
\newblock \href {https://doi.org/10.18653/v1/S19-2192} {{BUT}-{FIT} at
  {S}em{E}val-2019 task 7: Determining the rumour stance with pre-trained deep
  bidirectional transformers}.
\newblock In \emph{Proceedings of the 13th International Workshop on Semantic
  Evaluation}, pages 1097--1104, Minneapolis, Minnesota, USA. Association for
  Computational Linguistics.

\bibitem[{Fawcett(2006)}]{fawcett2006roc}
Tom Fawcett. 2006.
\newblock \href
  {https://doi.org/http://dx.doi.org/10.1016/j.patrec.2005.10.010} {{An
  introduction to ROC analysis}}.
\newblock \emph{Pattern Recognition Letters}, 27(8):861--874.

\bibitem[{Garc{\'\i}a-Cumbreras et~al.(2019)Garc{\'\i}a-Cumbreras,
  Jim{\'e}nez-Zafra, Montejo-R{\'a}ez, D{\'\i}az-Galiano, and
  Saquete}]{garcia-cumbreras-etal-2019-sinai}
Miguel~A. Garc{\'\i}a-Cumbreras, Salud~Mar{\'\i}a Jim{\'e}nez-Zafra, Arturo
  Montejo-R{\'a}ez, Manuel~Carlos D{\'\i}az-Galiano, and Estela Saquete. 2019.
\newblock \href {https://doi.org/10.18653/v1/S19-2196} {{SINAI}-{DL} at
  {S}em{E}val-2019 task 7: Data augmentation and temporal expressions}.
\newblock In \emph{Proceedings of the 13th International Workshop on Semantic
  Evaluation}, pages 1120--1124, Minneapolis, Minnesota, USA. Association for
  Computational Linguistics.

\bibitem[{Garc{\'\i}a~Lozano et~al.(2017)Garc{\'\i}a~Lozano, Lilja,
  Tj{\"o}rnhammar, and Karasalo}]{garcia-lozano-etal-2017-mama}
Marianela Garc{\'\i}a~Lozano, Hanna Lilja, Edward Tj{\"o}rnhammar, and Maja
  Karasalo. 2017.
\newblock \href {https://doi.org/10.18653/v1/S17-2084} {Mama edha at
  {S}em{E}val-2017 task 8: Stance classification with {CNN} and rules}.
\newblock In \emph{Proceedings of the 11th International Workshop on Semantic
  Evaluation ({S}em{E}val-2017)}, pages 481--485, Vancouver, Canada.
  Association for Computational Linguistics.

\bibitem[{Ghanem et~al.(2019)Ghanem, Cignarella, Bosco, Rosso, and
  Rangel~Pardo}]{ghanem-etal-2019-upv}
Bilal Ghanem, Alessandra~Teresa Cignarella, Cristina Bosco, Paolo Rosso, and
  Francisco~Manuel Rangel~Pardo. 2019.
\newblock \href {https://doi.org/10.18653/v1/S19-2197} {{UPV}-28-{UNITO} at
  {S}em{E}val-2019 task 7: Exploiting post{'}s nesting and syntax information
  for rumor stance classification}.
\newblock In \emph{Proceedings of the 13th International Workshop on Semantic
  Evaluation}, pages 1125--1131, Minneapolis, Minnesota, USA. Association for
  Computational Linguistics.

\bibitem[{Gorrell et~al.(2019)Gorrell, Kochkina, Liakata, Aker, Zubiaga,
  Bontcheva, and Derczynski}]{gorrell-etal-2019-semeval}
Genevieve Gorrell, Elena Kochkina, Maria Liakata, Ahmet Aker, Arkaitz Zubiaga,
  Kalina Bontcheva, and Leon Derczynski. 2019.
\newblock \href {https://doi.org/10.18653/v1/S19-2147} {{S}em{E}val-2019 task
  7: {R}umour{E}val, determining rumour veracity and support for rumours}.
\newblock In \emph{Proceedings of the 13th International Workshop on Semantic
  Evaluation}, pages 845--854, Minneapolis, Minnesota, USA. Association for
  Computational Linguistics.

\bibitem[{Hamidian and Diab(2019)}]{hamidian-diab-2019-gwu}
Sardar Hamidian and Mona Diab. 2019.
\newblock \href {https://doi.org/10.18653/v1/S19-2195} {{GWU} {NLP} at
  {S}em{E}val-2019 task 7: Hybrid pipeline for rumour veracity and stance
  classification on social media}.
\newblock In \emph{Proceedings of the 13th International Workshop on Semantic
  Evaluation}, pages 1115--1119, Minneapolis, Minnesota, USA. Association for
  Computational Linguistics.

\bibitem[{Huang and Ling(2005)}]{huang2005using}
Jin Huang and Charles~X Ling. 2005.
\newblock \href {https://ieeexplore.ieee.org/document/1388242} {{Using AUC and
  accuracy in evaluating learning algorithms}}.
\newblock \emph{IEEE Transactions on Knowledge and Data Engineering},
  17(3):299--310.

\bibitem[{Kochkina et~al.(2018)Kochkina, Liakata, and
  Zubiaga}]{kochkina-etal-2018-one}
Elena Kochkina, Maria Liakata, and Arkaitz Zubiaga. 2018.
\newblock \href {https://www.aclweb.org/anthology/C18-1288} {All-in-one:
  Multi-task learning for rumour verification}.
\newblock In \emph{Proceedings of the 27th International Conference on
  Computational Linguistics}, pages 3402--3413, Santa Fe, New Mexico, USA.
  Association for Computational Linguistics.

\bibitem[{Kumar and Carley(2019)}]{kumar-carley-2019-tree}
Sumeet Kumar and Kathleen Carley. 2019.
\newblock \href {https://doi.org/10.18653/v1/P19-1498} {Tree {LSTM}s with
  convolution units to predict stance and rumor veracity in social media
  conversations}.
\newblock In \emph{Proceedings of the 57th Annual Meeting of the Association
  for Computational Linguistics}, pages 5047--5058, Florence, Italy.
  Association for Computational Linguistics.

\bibitem[{Li et~al.(2019{\natexlab{a}})Li, Zhang, and
  Si}]{li-etal-2019-eventai}
Quanzhi Li, Qiong Zhang, and Luo Si. 2019{\natexlab{a}}.
\newblock \href {https://doi.org/10.18653/v1/S19-2148} {event{AI} at
  {S}em{E}val-2019 task 7: Rumor detection on social media by exploiting
  content, user credibility and propagation information}.
\newblock In \emph{Proceedings of the 13th International Workshop on Semantic
  Evaluation}, pages 855--859, Minneapolis, Minnesota, USA. Association for
  Computational Linguistics.

\bibitem[{Li et~al.(2019{\natexlab{b}})Li, Zhang, and Si}]{li-etal-2019-rumor}
Quanzhi Li, Qiong Zhang, and Luo Si. 2019{\natexlab{b}}.
\newblock \href {https://doi.org/10.18653/v1/P19-1113} {Rumor detection by
  exploiting user credibility information, attention and multi-task learning}.
\newblock In \emph{Proceedings of the 57th Annual Meeting of the Association
  for Computational Linguistics}, pages 1173--1179, Florence, Italy.
  Association for Computational Linguistics.

\bibitem[{Liu et~al.(2019)Liu, Goel, Yelahanka~Raghuprasad, and
  Muresan}]{liu-etal-2019-columbia}
Zhuoran Liu, Shivali Goel, Mukund Yelahanka~Raghuprasad, and Smaranda Muresan.
  2019.
\newblock \href {https://doi.org/10.18653/v1/S19-2194} {{C}olumbia at
  {S}em{E}val-2019 task 7: Multi-task learning for stance classification and
  rumour verification}.
\newblock In \emph{Proceedings of the 13th International Workshop on Semantic
  Evaluation}, pages 1110--1114, Minneapolis, Minnesota, USA. Association for
  Computational Linguistics.

\bibitem[{Mendoza et~al.(2010)Mendoza, Poblete, and
  Castillo}]{mendoza2010twitter}
Marcelo Mendoza, Barbara Poblete, and Carlos Castillo. 2010.
\newblock \href {http://snap.stanford.edu/soma2010/papers/soma2010_11.pdf}
  {{Twitter under crisis: can we trust what we RT?}}
\newblock In \emph{Proceedings of the First Workshop on Social Media
  Analytics}, pages 71--79, Washington, DC, USA. Association for Computing
  Machinery.

\bibitem[{Mohammad et~al.(2016)Mohammad, Kiritchenko, Sobhani, Zhu, and
  Cherry}]{mohammad-etal-2016-semeval}
Saif Mohammad, Svetlana Kiritchenko, Parinaz Sobhani, Xiaodan Zhu, and Colin
  Cherry. 2016.
\newblock \href {https://doi.org/10.18653/v1/S16-1003} {{S}em{E}val-2016 task
  6: Detecting stance in tweets}.
\newblock In \emph{Proceedings of the 10th International Workshop on Semantic
  Evaluation ({S}em{E}val-2016)}, pages 31--41, San Diego, California.
  Association for Computational Linguistics.

\bibitem[{Singh et~al.(2017)Singh, Narayan, Akhtar, Ekbal, and
  Bhattacharyya}]{singh-etal-2017-iitp}
Vikram Singh, Sunny Narayan, Md~Shad Akhtar, Asif Ekbal, and Pushpak
  Bhattacharyya. 2017.
\newblock \href {https://doi.org/10.18653/v1/S17-2087} {{IITP} at
  {S}em{E}val-2017 task 8 : A supervised approach for rumour evaluation}.
\newblock In \emph{Proceedings of the 11th International Workshop on Semantic
  Evaluation ({S}em{E}val-2017)}, pages 497--501, Vancouver, Canada.
  Association for Computational Linguistics.

\bibitem[{Srivastava et~al.(2017)Srivastava, Rehm, and
  Moreno~Schneider}]{srivastava-etal-2017-dfki}
Ankit Srivastava, Georg Rehm, and Julian Moreno~Schneider. 2017.
\newblock \href {https://doi.org/10.18653/v1/S17-2085} {{DFKI}-{DKT} at
  {S}em{E}val-2017 task 8: Rumour detection and classification using cascading
  heuristics}.
\newblock In \emph{Proceedings of the 11th International Workshop on Semantic
  Evaluation ({S}em{E}val-2017)}, pages 486--490, Vancouver, Canada.
  Association for Computational Linguistics.

\bibitem[{Wang et~al.(2017)Wang, Lan, and Wu}]{wang-etal-2017-ecnu}
Feixiang Wang, Man Lan, and Yuanbin Wu. 2017.
\newblock \href {https://doi.org/10.18653/v1/S17-2086} {{ECNU} at
  {S}em{E}val-2017 task 8: Rumour evaluation using effective features and
  supervised ensemble models}.
\newblock In \emph{Proceedings of the 11th International Workshop on Semantic
  Evaluation ({S}em{E}val-2017)}, pages 491--496, Vancouver, Canada.
  Association for Computational Linguistics.

\bibitem[{Yang et~al.(2019)Yang, Xie, Liu, and Yu}]{yang-etal-2019-blcu}
Ruoyao Yang, Wanying Xie, Chunhua Liu, and Dong Yu. 2019.
\newblock \href {https://doi.org/10.18653/v1/S19-2191} {{BLCU}{\_}{NLP} at
  {S}em{E}val-2019 task 7: An inference chain-based {GPT} model for rumour
  evaluation}.
\newblock In \emph{Proceedings of the 13th International Workshop on Semantic
  Evaluation}, pages 1090--1096, Minneapolis, Minnesota, USA. Association for
  Computational Linguistics.

\bibitem[{Yijing et~al.(2016)Yijing, Haixiang, Xiao, Yanan, and
  Jinling}]{yijing-etal-2016}
Li~Yijing, Guo Haixiang, Liu Xiao, Li~Yanan, and Li~Jinling. 2016.
\newblock \href {https://doi.org/https://doi.org/10.1016/j.knosys.2015.11.013}
  {Adapted ensemble classification algorithm based on multiple classifier
  system and feature selection for classifying multi-class imbalanced data}.
\newblock \emph{Knowledge-Based Systems}, 94:88--104.

\bibitem[{Zubiaga et~al.(2018)Zubiaga, Aker, Bontcheva, Liakata, and
  Procter}]{zubiaga-etal-2018}
Arkaitz Zubiaga, Ahmet Aker, Kalina Bontcheva, Maria Liakata, and Rob Procter.
  2018.
\newblock \href {https://doi.org/10.1145/3161603} {Detection and resolution of
  rumours in social media: A survey}.
\newblock \emph{ACM Computing Surveys}, 51(2):32:1--32:36.

\bibitem[{Zubiaga et~al.(2016)Zubiaga, Liakata, Procter, Hoi, and
  Tolmie}]{zubiaga-etal-2016-plos}
Arkaitz Zubiaga, Maria Liakata, Rob Procter, Geraldine Wong~Sak Hoi, and Peter
  Tolmie. 2016.
\newblock \href {https://doi.org/10.1371/journal.pone.0150989} {Analysing how
  people orient to and spread rumours in social media by looking at
  conversational threads}.
\newblock \emph{Plos One}, 11(3).

\end{thebibliography}
\bibliographystyle{acl_natbib}

\appendix

\section{Confusion matrices for all RumourEval 2019 systems}
\label{sec:appendix}
\begin{figure*}[!t]
\begin{center}
  \includegraphics[width=1.\textwidth]{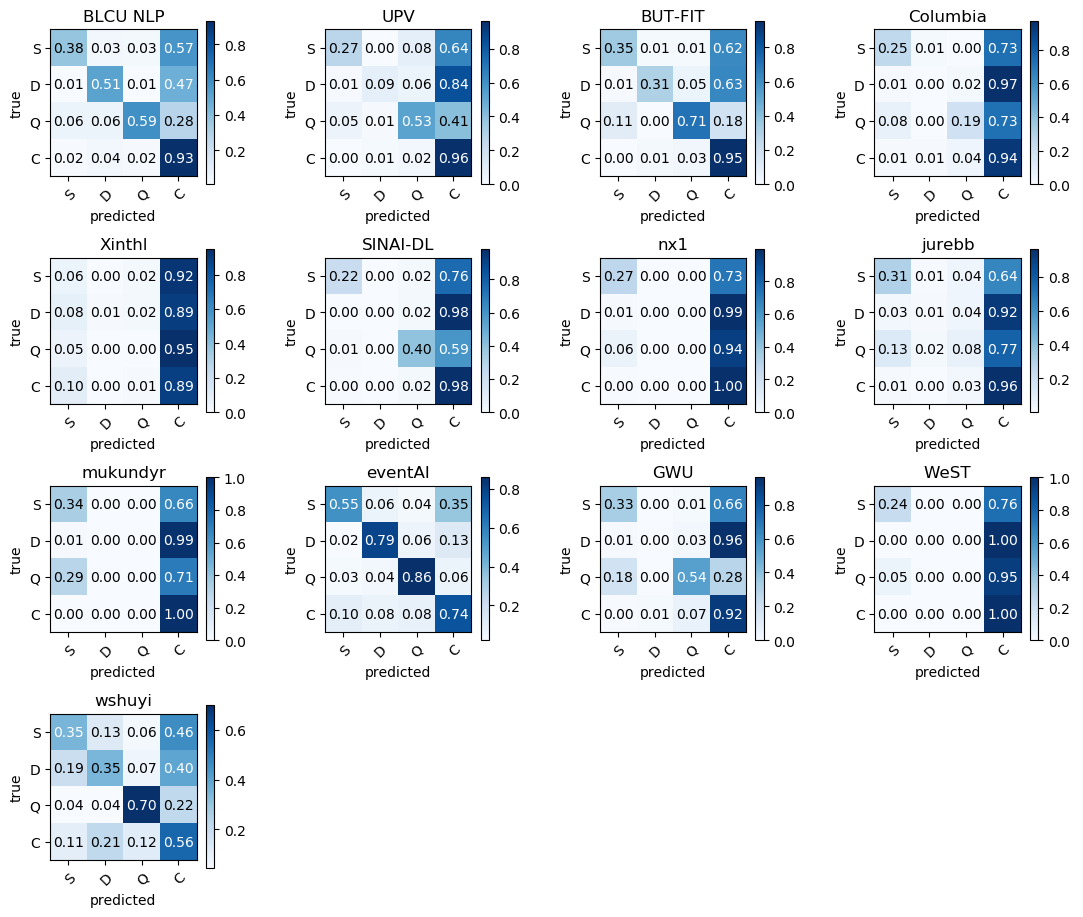}
  \caption{Confusion matrix for all systems from RumourEval 2019.}
  \label{fig:cm2019_all}
\end{center}
\end{figure*}

For completeness, Figure \ref{fig:cm2019_all} shows the confusion matrices of all systems submitted to RumourEval 2019. Apart from the four systems discussed in Section \ref{sec:results}, all other systems fails to correctly classify the large majority of \textit{deny} instances.

\end{document}